\documentclass[runningheads]{llncs}
\usepackage{graphicx}

\usepackage{hyperref}
\hypersetup{
    colorlinks=true,
    linkcolor=blue,
    filecolor=magenta,      
    urlcolor=black,
}

\begin{document}

\title{Augmented Data as an Auxiliary Plug-in Towards Categorization of Crowdsourced Heritage Data}
\titlerunning{Categorization of Crowdsourced Heritage Data}
\author{Shashidhar Veerappa Kudari \and
Akshaykumar Gunari \and
Adarsh Jamadandi \and
Ramesh Ashok Tabib \and
Uma Mudenagudi
}

\authorrunning{Shashidhar et al.}
%
\institute{KLE Technological University, Hubballi\\
shashidharvk100@gmail.com, akshaygunari@gmail.com, adarsh.cto@tweaklabsinc.com, ramesh\_t@kletech.ac.in, uma@kletech.ac.in 
}
\maketitle              
\begin{abstract}
In this paper, we propose a strategy to mitigate the problem of inefficient clustering performance by introducing data augmentation as an auxiliary plug-in. Classical clustering techniques such as K-means, Gaussian mixture model and spectral clustering are central to many data-driven applications. However, recently unsupervised simultaneous feature learning and clustering using neural networks also known as Deep Embedded Clustering (DEC) has gained prominence. Pioneering works on deep feature clustering focus on defining relevant clustering loss function and choosing the right neural network for extracting features. A central problem in all these cases is data sparsity accompanied by high intra-class and low inter-class variance, which subsequently leads to poor clustering performance and erroneous candidate assignments. Towards this, we employ data augmentation techniques to improve the density of the clusters, thus improving the overall performance. We train a variant of Convolutional Autoencoder (CAE) with augmented data to construct the initial feature space as a novel model for deep clustering. We demonstrate the results of proposed strategy on crowdsourced Indian Heritage dataset. Extensive experiments show consistent improvements over existing works.

\end{abstract}

\section{Introduction}

In this paper, we propose a novel training mechanism to mitigate problems such as data sparsity, high inter-class variance, and low intra-class variance which leads to poor clustering performance. Traditional clustering algorithms such as K-means, Gaussian mixture models (GMM) \cite{paper18}, and spectral clustering \cite{Paper19} rely largely on the notion of \emph{distance}; for example,  K-means \cite{Paper20} uses Euclidean distance to assign data points to clusters.  Recent advances in deep learning has led to emergence of clustering techniques parameterized by deep neural networks \cite{Paper1} \cite{Paper7} \cite{Paper13} \cite{Paper14} \cite{Paper15} attempting to jointly learn representations, and perform clustering relying on tools like Stochastic Gradient Descent and backpropagation with a clustering objective function. This introduces  challenges in choosing an appropriate neural network architecture, and a right clustering objective function. Recent methods \cite{Paper2}\cite{Paper5}, attempt to circumvent these problems, limited literature show investigations on the effect of data sparsity and high intra-class variance, usually found in crowdsourced  cultural heritage datasets. The apparent architectural differences arise due to data acquisition methods and cultural similarities might lead to assignment of false  clusters. In this paper, we empirically demonstrate the use of different transformations such as random - scaling, rotation and shearing as data augmentation techniques towards increasing the data density, yielding superior clustering performance. 

Crowd-sourcing facilitates desired data at scale and involves task owners relying on a large batch of supposedly anonymous human resources with varying expertise contributing a diversified amount of data. In our case, we are interested in obtaining a large image corpus of Indian Heritage Sites with the hindsight of large scale 3D reconstruction towards digital archival and preservation.  An essential step in this pipeline is to formulate an efficient deep clustering method towards  mitigate issues outlined above. Towards this -

\begin{itemize}
	\item We propose a novel training strategy to circumvent the problem of poor clustering performance by
	
	\begin{itemize}
		\item introducing data augmentation as an auxiliary plug-in for deep embedded clustering
		\begin{itemize}
		    \item to densify data and facilitate better feature representation considering limited data. 
		    \item to address data with high intra-class and low inter-class variance.
		    \item to augment data using affine transforms (rotation, scaling and shearing).
		\end{itemize}
	        \item incorporating Consistency Constraint Loss (CCL) with Mean Squared Error (MSE) Loss to handle introduced transformations.
	\end{itemize}

\item We demonstrate our proposed strategy on a crowdourced Indian heritage dataset and show consistent improvements over existing works.
\end{itemize}

In Section \ref{RW}, we discuss contemporary works related to clustering. In Section \ref{M}, we propose a strategy to circumvent the problem of poor clustering performance. In Section \ref{Exp}, we discuss experimental setup carried out on Indian Heritage Dataset. In Section \ref{RD}, we demonstrate results through quantitative and qualitative metrics, and conclude in Section \ref{Con}.

\newpage

\section{Related Works} 
\label{RW}
In this section, we discusses contemporary works addressing clustering using deep features. Classical clustering techniques such as K-means \cite{Paper20}, Gaussian Mixture Models (GMM) \cite{paper18}, and Spectral clustering \cite{Paper19} are limited by their distance metrics and perform poorly when the dimensionality is high. Towards this, recent techniques such as Deep Embedded Clustering (DEC) \cite{Paper2}, Improved Deep Embedded Clustering \cite{Paper5} extract deep features towards categorization in lower dimension embedding space. 

Recent advances in deep neural networks have ushered a strategy of parameterizing clustering algorithms with neural networks. Deep Embedded Clustering (DEC), proposed by authors in \cite{Paper2}, pioneered the idea of using deep neural networks to learn representations and solve for cluster assignment jointly. The method involves using Stochastic Gradient Descent coupled with backpropagation to extract deep features while simultaneously learning the underlying representations. 
However, as authors in \cite{Paper5} point, the choice of clustering loss tends to distort the feature space, which consequently affects the overall clustering performance. To mitigate this, the authors propose an under-complete autoencoder to preserve the data structure, leading to improved clustering performance. 
Inspired by these works, we propose a method to improve Clustering performance by densifying the data distribution. We hypothesize that data distribution sparsity as a significant deterrent in clustering. The problem is further exacerbated when the data exhibits high intra-class variance.  We empirically show that using data augmentation as an auxiliary plug-in helps in improving cluster performance. Extensive experiments on cultural heritage dataset shows consistent improvements over existing methods.

\section{Categorization of Crowdsourced Heritage Data}
\label{M}

The crowdsourced heritage data arrives in the incremental fashion, where the number of classes and number of images belonging to class are obscure. More likely, we observe that images belonging to a particular class may arrive in large number while very few samples may arrive for some other classes. This brings the problems of class imbalance and data sparsity. Due to data sparsity, deep learning techniques used for the feature representation of the images fail in their task, making the clustering performance poor. Deep learning architectures like Convolutional Autoencoders (CAE) are sensitive to these problems. Towards this, we attempt to  mitigate the data sparsity issue via data augmentation.

\begin{figure}
\begin{center}
\includegraphics[width= 1.0 \linewidth]{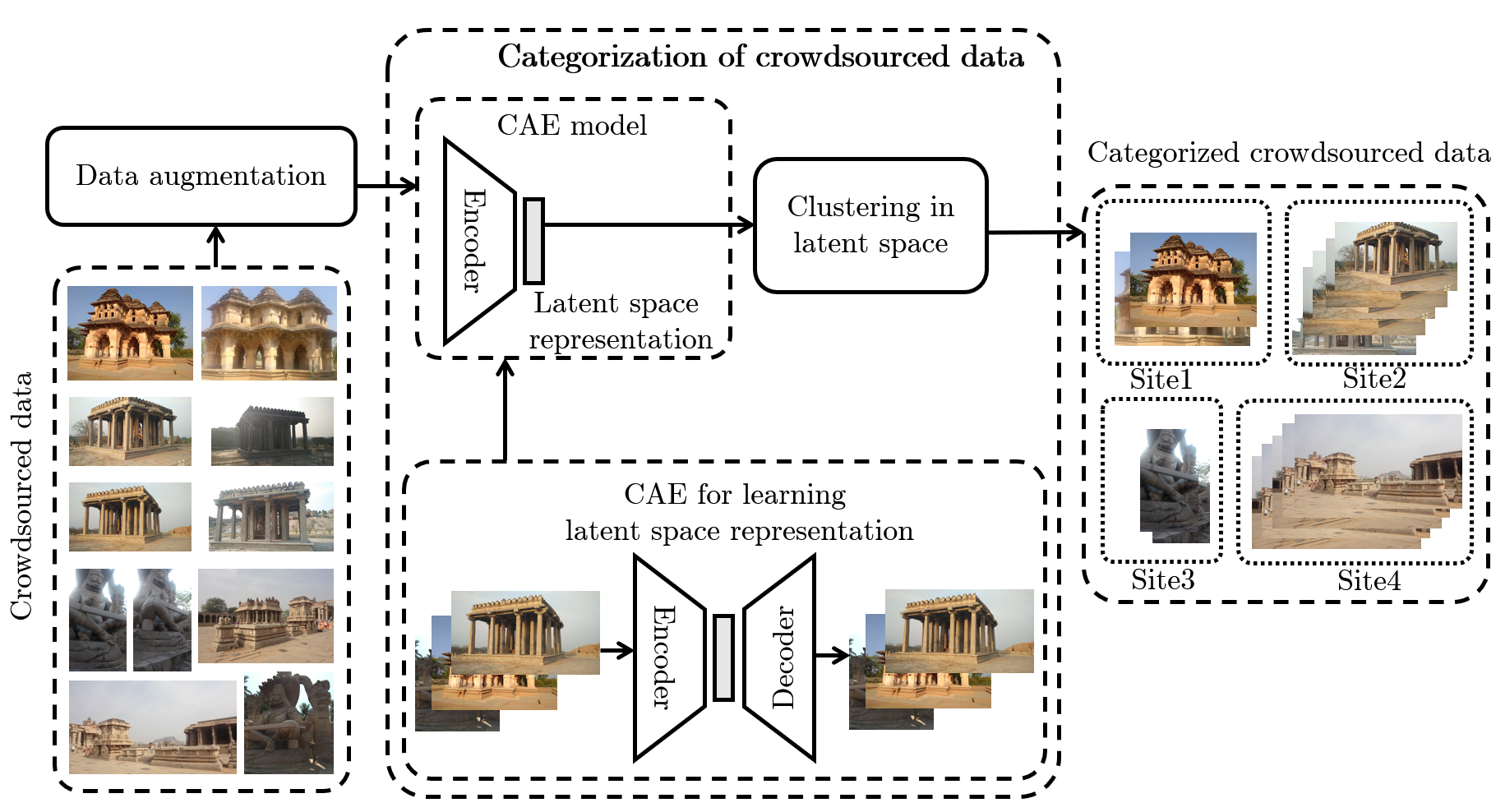}
\caption{Categorization of crowdsourced heritage data}
\label{fig:four}
\end{center}
\end{figure}

We increase the density of the data, by performing three kinds of data transformations, i.e random rotation, random shear and random scaling on the original data, as this would generally make the model more robust in terms of learning. These techniques tend to provide more generic and genuine data. There are many other augmentation techniques as described in  \cite{Paper4}\cite{Paper6}\cite{Paper8}\cite{Paper12}  used to increase the data density and class imbalance \cite{Paper3}\cite{Paper9}.

Convolutional Autoencoder (CAE) have proven to be effective in case of  classification, clustering and object detection. We combine the augmented data with original data to train CAE to generate embeddings towards clustering.

Considering $x_i$, $i \in \{1,...,m\}$ as an image, the transformation $t_j$, $j \in \{1,.., s\}$ applied on $x_i$ generates transformed image $x_i^{t_s}$, represented as $x_i^{t} = T(x_i)$. 
The total number of images after augmentation are $N$, where $N = s\times m$.

The traditional objective function used for training the CAE is Mean Squared Error(MSE) between the input $x_i$ and decoder output $x_i^{'}$ which is as:
\begin{equation}
\label{equation:1}
         MSE(x, x^`) = \frac{\sum_{i=1}^{N}{(x_i-x_i^`)^2}}{N}
\end{equation}

The MSE loss for training CAE limits information about the relation between original data and the augmented data.
To overcome this, we incorporate the Consistency Constraints \cite{Paper17}. The Consistency Constraints are seen to be effective in the Semi Supervised Learning (SSL). A Consistency Constraint Loss (CCL) can be incorporated by enforcing the predictions of a data sample and its transformed counterpart (which can be obtained by randomly rotating, shearing or scaling the images) to be minimal. The CCL Loss  is defined as follows:


\begin{equation}
        \label{eq:2}
        L_{CCL} = \frac{1}{NK}\sum_{i=1}^{N}\sum_{k=1}^{K}\parallel p(k|i) - p^{t}(k|i)\parallel
\end{equation}

where $N$ represents the total number of data points and the $K$ represents total number of
clusters, $p(k|i)$ represents the probability of assignment of each image
$x_i$, and $p^{t}(k|i)$ represents the probability of assignment of randomly transformed image $x_i^{t}$
to cluster $k$. $p(k|i)$ is parameterized by assuming they follow the 
Student's T distribution as follows:
\begin{equation}
        p(k|i) 	\propto (1+\frac{\parallel z_i + \mu_k \parallel^2}{\alpha})^{-\frac{\alpha+1}{2}}
\end{equation}

Here $z_i$ is the feature representation of the image $x_i$, $\mu_k$ represents the cluster center of cluster $k$. If $U$ represents the cluster centers then $U = \{ \mu_k,k = 1....K\}$ which are initialized by K-means and tuned as the training progress.
\newline
The overall objective function of CAE is now defined as:
\begin{equation}
    Loss = MSE(x, x') + L_{CCL}
\end{equation}
We use K-means technique to quantify our results, and depict how augmentation can improve the performance of the clustering. We show how 
data augmentation can improve the performance of the existing state of art methods Deep Embedded Clustering (DEC) and Improved Deep Embedded Clustering (IDEC), where CAE is used as the initial feature extractor. We provide the extensive ablation study of these methods over the combinations of different CAEs trained.

\section{Experiments}
\label{Exp}

\subsection{Dataset}
We extensively experiment on crowdsourced Indian Digital Heritage (IDH) Dataset. The dataset is collected through a platform sourced by crowd. We consider 10  classes of this dataset with 150 images per class towards experimentation as these 10 classes consists of high intra-class and low inter-class variance. The considered dataset undergoes augmentation like random rotation, random shearing and random scaling. Random rotation of images is performed over the range of 0-90 degrees, random shear is performed over 50 degrees of transformation intensity and random scaling is performed over the scale of 0.5 - 1.0. We generate around 6000 images through these transformations. The same dataset is used throughout the experimentation to maintain the uniformity in comparison of results in different experiments in different environments.

\subsection{Training Setup}

\begin{itemize}
    \item {Runtime Environment: Nvidia GP107CL Quadro P620}

    \item {Architecture: Autoencoder}
    \begin{itemize}
    
    \item {Encoder: }
    \begin{itemize}
        \item {Contains 4 VGG Blocks}
        \item {VGG Block has 2 Convolution Layers followed by a maxpooling layer} 
        \item {Batch normalization layer was used at the end of each layer before the activation function}
        \item {Activation Function: ReLU }
    \end{itemize}
    \item {Decoder: }
  \begin{itemize}
        \item {Decoder part of the model consists of convolution transpose layers  with batch normalization layer at the end of each layer before the activation function}
        \item {Activation Function: ReLU, Sigmoid (Output Layer)} 
    \end{itemize}
    \end{itemize}

    \item {Batch Size: 16}
    \item {Learning Rate: 0.001}
    \item {Number of Epochs: 500 (CAE), 2000 (DEC and IDEC)}
    \item {Optimizer: Adam}
\end{itemize}

\subsection{Evaluation Metrics}
Towards evaluation of proposed strategy and comparision with state of the art methods, we use Unsupervised Clustering Accuracy (ACC), Normalized Mutual Information (NMI), Adjusted Rand Index (ARI).
\subsubsection{Unsupervised Clustering Accuracy (ACC):}
It uses a mapping function $m$ to find the best mapping between the cluster assignment output $c$ of the algorithm with the ground truth $y$ which can be defined as:

\begin{equation}
    ACC = max_m \frac{\sum_{i=1}^{N} 1\{y_i = m(c_i)\}}{N}
\end{equation}

\subsubsection{Normalized Mutual Information (NMI):}
It measures the mutual information $I(y,c)$ between the cluster assignments$ c$ and the ground truth labels $y$ and is normalized by the average of entropy of both ground labels $H(y)$ and the cluster assignments $H(c)$, and can be defined as:

\begin{equation}
    NMI = \frac{I(y,c)}{\frac{1}{2}[H(y) + H(c)]}
\end{equation}

\subsubsection{Adjusted Rand Index (ARI):}
It computes a similarity measure between two clusterings by considering all pairs of samples and counting pairs that are assigned in the same or different clusters in the predicted and true clusterings. It is defined as:

\begin{equation}
    ARI = \frac{Index - ExpectedIndex}{MaxIndex - ExpectedIndex}
\end{equation}

\section{Results And Discussions}
\label{RD}
In this section, we discuss the results of proposed strategy towards categorization of crowdsourced Indian Heritage (IDH) dataset and compare the results with state of the art methods.
\hspace{13px}

We measure the clustering performance by reporting the unsupervised clustering accuracy, NMI and ARI. From Table \ref{tab:my-table1} we observe, CAE trained with data augmentation yields better performance over CAE trained without augmention. We see an improvement of \textbf{2.74\%} when trained with MSE loss and an improvement of \textbf{15.21\%} when trained with a combination of MSE and CCL, as CCL loss mainly depends on augmented data. This improvement is significant in context of clustering data with high intra-class variance.

\begin{table}[]
\centering
\caption{Performance of CAE trained with and without augmented data. CAE-WAug represents CAE model trained without augmented data and CAE-Aug represents CAE model trained with augmented data.}
\label{tab:my-table1}
\begin{tabular}{l|cll|lll}
\hline
                                                    & \multicolumn{3}{c|}{MSE} & \multicolumn{3}{c|}{MSE + CCL} \\ \hline
\multicolumn{1}{c|}{}                               & ACC    & NMI    & ARI    & ACC      & NMI      & ARI      \\ \hline
\begin{tabular}[c]{@{}l@{}}CAE-WAug\end{tabular} & 0.5424 & 0.4880 & 0.3377 & 0.4035   & 0.3752   & 0.1880   \\
\begin{tabular}[c]{@{}l@{}}CAE-Aug\end{tabular}  & \textbf{0.5698} & \textbf{0.5327} & \textbf{0.4000} & \textbf{0.5566}   & \textbf{0.4936}   & \textbf{0.3336}   \\ \hline
\end{tabular}
\end{table}

To discern the effect of individual augmentation techniques (rotate, scale and shear), we choose samples in the combination of - \{ori, rot\}, \{ori, sher\} and \{ori, scal\}. The results are presented in Table \ref{tab:my-table2}. We observe, set \{ori, rot\} shows poor performance compared to augmentations consisting of \{ori, sher\} and \{ori, scal\}. We hypothesize that, the performance drop for \{ori, sher\} can be attributed to the fact that, the CAE is not equipped with appropriate symmetry inductive bias that enables it to learn rotation-invariant features.

\begin{table}[]
\centering
\caption{Effect of augmentation on performance of the original data. Original, rotated, sheared, scaled data are represented as ori, rot, sher and scal respectively.}
\label{tab:my-table2}
\begin{tabular}{l|lll|lll}
\hline
                                                              & \multicolumn{3}{c|}{CAE - MSE} & \multicolumn{3}{c|}{CAE - (MSE + CCL)}     \\ \hline
\multicolumn{1}{c|}{}                                         & ACC      & NMI      & ARI      & ACC    & NMI    & \multicolumn{1}{l|}{ARI} \\ \hline
\begin{tabular}[c]{@{}l@{}}ori  + rot\end{tabular} & 0.4355   & 0.3168   & 0.2155   & 0.3491 & 0.2681 & 0.1493                   \\
                                                              &          &          &          &        &        &                          \\
\begin{tabular}[c]{@{}l@{}}ori  + sher\end{tabular} & 0.5189   & 0.4480   & 0.3144   & 0.4413 & 0.3574 & 0.2091                   \\
                                                              &          &          &          &        &        &                          \\
\begin{tabular}[c]{@{}l@{}}ori  + scal\end{tabular}  & 0.5183   & 0.4168   & 0.2846   & 0.4346 & 0.3419 & 0.2151                   \\ \hline
\end{tabular}
\end{table}

\subsection{Ablation Study}

In this section, we perform ablation study using DEC \cite{Paper2} and IDEC \cite{Paper5} with and without considering augmentation. 
In Table \ref{tab:my-table3}, we provide the ablation study of DEC which is unsupervised clustering technique that joinlty optimizes the cluster centers and the parameters of the CAE. KL-divergence between the auxiliary and target distribution optimizes the objective function. From Table \ref{tab:my-table3} we infer, providing CAE with the augmented data followed by DEC considering original data increases the accuracy by \textbf{5.61\%}. While providing augmented data to DEC, with CAE being trained with original data hinders the performance. Hence, only CAE is trained with with original and augmented data ensuring the objective is met.

\begin{table}[]
\centering
\caption{Comparing  results  of  proposed  methodology  with  DEC \cite{Paper2}.  CAE-WAugand  CAE-Aug  refers  to  CAE  trained  without  and  with  augmentation  respectively.DEC-WAug  and  DEC-Aug  refers  to  DEC  trained  without  and  with  augmnetationrespectively.  We  show  how  combination  of  augmentation  applied  to  CAE  and  DECmay affect the clustering performance.}
\label{tab:my-table3}
\resizebox{\textwidth}{!}{%
\begin{tabular}{llllllllll}
\cline{1-7}
\multicolumn{1}{|l|}{Loss→}            & \multicolumn{3}{l|}{MSE}             & \multicolumn{3}{l|}{MSE + CCL}       &  &  &  \\ \cline{1-7}
\multicolumn{1}{|l|}{Method↓, Metric→} & ACC & NMI & \multicolumn{1}{l|}{ARI} & ACC & NMI & \multicolumn{1}{l|}{ARI} &  &  &  \\ \cline{1-7}
\multicolumn{1}{|l|}{\begin{tabular}[c]{@{}l@{}}DEC \cite{Paper2} \\ CAE - WAug\\ + DEC - WAug\end{tabular}} &
  0.4113 &
  0.3874 &
  \multicolumn{1}{l|}{0.2271} &
  0.3625 &
  0.3625 &
  \multicolumn{1}{l|}{0.8196} &
   &
   &
   \\ \cline{1-7}
\multicolumn{1}{|l|}{\begin{tabular}[c]{@{}l@{}}CAE - WAug\\ + DEC Aug\end{tabular}} &
  0.3096 &
  0.3013 &
  \multicolumn{1}{l|}{0.1530} &
  0.2927 &
  0.2210 &
  \multicolumn{1}{l|}{0.1201} &
   &
   &
   \\ \cline{1-7}
\multicolumn{1}{|l|}{\begin{tabular}[c]{@{}l@{}}CAE - Aug\\ + DEC WAug\end{tabular}} &
  \textbf{0.4674} &
  \textbf{0.4876} &
  \multicolumn{1}{l|}{\textbf{0.3076}} &
  \textbf{0.4492} &
  \textbf{0.4665} &
  \multicolumn{1}{l|}{\textbf{0.2716}} &
   &
   &
   \\ \cline{1-7}
                                       &     &     &                          &     &     &                          &  &  &  \\
                                       &     &     &                          &     &     &                          &  &  &  \\
                                       &     &     &                          &     &     &                          &  &  &  \\
                                       &     &     &                          &     &     &                          &  &  &  \\
                                       &     &     &                          &     &     &                          &  &  & 
\end{tabular}%
}
\end{table}

In Table. \ref{tab:my-table4} we provide the ablation study of the Improved Deep Embedded Clustering (IDEC).  IDEC is an improvemnet over IDEC, which not only jointly optimize the cluster centers and parameters of the CAE, but also preserve the local structure information. They use KL divergence between the auxiliary and target distribution as their objective function along with the MSE loss of the CAE. From Table \ref{tab:my-table4} it can be observed that providing CAE with augmentated data with MSE+CCL loss, then providing the trained CAE to IDEC, where IDEC is trained on original data improves the performance by depicting the increase in accuracy by \textbf{7.43\%}. While training the IDEC with augmented data with CAE trained on original data only hinders the performance.

\begin{table}[]
\centering
\caption{Comparing results of proposed methodology with IDEC \cite{Paper5}. CAE-WAug and CAE-Aug refers to CAE trained without and with augmented data respectively. IDEC-WAug and IDEC-Aug refers to IDEC trained without and with augmneted data respectively. We show how combination of augmentation applied to CAE and IDEC may affect the clustering performance.}
\label{tab:my-table4}
\resizebox{\textwidth}{!}{%
\begin{tabular}{llllllllll}
\cline{1-7}
\multicolumn{1}{|l|}{Loss→}            & \multicolumn{3}{l|}{MSE}             & \multicolumn{3}{l|}{MSE + CCL}       &  &  &  \\ \cline{1-7}
\multicolumn{1}{|l|}{Method↓, Metric→} & ACC & NMI & \multicolumn{1}{l|}{ARI} & ACC & NMI & \multicolumn{1}{l|}{ARI} &  &  &  \\ \cline{1-7}
\multicolumn{1}{|l|}{\begin{tabular}[c]{@{}l@{}}IDEC \cite{Paper5}\\ CAE - WAug\\ + IDEC - WAug\end{tabular}} &
  0.5098 &
  0.4903 &
  \multicolumn{1}{l|}{0.3020} &
  0.3625 &
  0.4340 &
  \multicolumn{1}{l|}{0.2511} &
   &
   &
   \\ \cline{1-7}
\multicolumn{1}{|l|}{\begin{tabular}[c]{@{}l@{}}CAE - WAug\\ + IDEC Aug\end{tabular}} &
  0.3611 &
  0.3300 &
  \multicolumn{1}{l|}{0.1703} &
  0.3514 &
  0.3335 &
  \multicolumn{1}{l|}{0.1856} &
   &
   &
   \\ \cline{1-7}
\multicolumn{1}{|l|}{\begin{tabular}[c]{@{}l@{}}CAE - Aug\\ + IDEC WAug\end{tabular}} &
  \textbf{0.4983} &
  \textbf{0.4942} &
  \multicolumn{1}{l|}{\textbf{0.3130}} &
  \textbf{0.5841} &
  \textbf{0.4340} &
  \multicolumn{1}{l|}{\textbf{0.3662}} &
   &
   &
   \\ \cline{1-7}
                                       &     &     &                          &     &     &                          &  &  &  \\
                                       &     &     &                          &     &     &                          &  &  &  \\
                                       &     &     &                          &     &     &                          &  &  &  \\
                                       &     &     &                          &     &     &                          &  &  &  \\
                                       &     &     &                          &     &     &                          &  &  & 
\end{tabular}%
}
\end{table}

From the experiments we observe, it is better to train the CAE with MSE+CCL as the integrity loss, with augmented data. The CAE trained in such environment is incorporated for initial feature representation to the IDEC by providing original data, to perform better than other methods.

\section{Conclusions}
\label{Con}
In this paper, we have defined data augmentation as an auxiliary plug-in for deep embedded clustering, that densifies data helping in accurate clustering performance. We have demonstrated how data augmentation helps to increase the data density yielding superior clustering performance when the data is considerably less in amount. Extensive experimentation is done on setting up the right objective function.  Our main objective is to cluster data with very less inter-class variance and very high intra-class variance. We have demonstrated our experiments on crowdsourced heritage dataset. We also show, how certain augmentation techniques uphold the clustering objectives (such as random shear and random scale), while some of them hinders the same (random rotation).

\bibliographystyle{splncs04}
\bibliography{Biblography}

\end{document}